\documentclass{article} 
\usepackage{iclr2027_conference,times}
\iclrfinalcopy

\usepackage{amsmath,amsfonts,bm}

\def\eqref#1{equation~\ref{#1}}

\def\1{\bm{1}}

\DeclareMathAlphabet{\mathsfit}{\encodingdefault}{\sfdefault}{m}{sl}
\SetMathAlphabet{\mathsfit}{bold}{\encodingdefault}{\sfdefault}{bx}{n}

\usepackage{hyperref}
\usepackage{url}
\usepackage{booktabs}
\usepackage{multirow}
\usepackage{adjustbox}
\usepackage{graphicx}
\usepackage{colortbl}
\usepackage{amssymb}

\definecolor{TableHeader}{RGB}{245,247,250}
\definecolor{TableGroup}{RGB}{247,248,250}
\definecolor{LoCoBlue}{RGB}{235,243,255}
\definecolor{LoCoText}{RGB}{40,86,158}

\newcommand{\ourcell}[1]{\cellcolor{LoCoBlue}#1}
\newcommand{\ourlabel}{\multirow[t]{2}{*}{\cellcolor{LoCoBlue}\textbf{\textcolor{LoCoText}{LoCo (Ours)}}}}
\newcommand{\ourblank}{\cellcolor{LoCoBlue}}

\newcommand{\blfootnote}[1]{%
  \begingroup
  \renewcommand{\thefootnote}{}%
  \footnotetext{#1}%
  \addtocounter{footnote}{-1}%
  \endgroup
}
\newcommand{\cbox}[1]{\makebox[\dimexpr\linewidth-2\tabcolsep\relax][c]{#1}}

\title{Think Less, Locally: Local Preference Learning for Reasoning Compression}

\author{%
    \cbox{Muyao Yuan\textsuperscript{1*}~~~Liyan Xu\textsuperscript{2*}$^{\ddagger}$~~~Weizhan Zhang\textsuperscript{1\textdagger}}
    \\[0.08cm]
    \cbox{\textbf{Muyan Jiao\textsuperscript{1}}~~~\textbf{Jiangnan Li\textsuperscript{2}}~~~\textbf{Jinchao Zhang\textsuperscript{2}}~~~\textbf{Jie Zhou\textsuperscript{2\textdagger}}}
    \\[0.15cm]
    \cbox{$^{1}$MOEKLINNS, School of Computer Science and Technology, Xi'an Jiaotong University}
    \\[0.03cm]
    \cbox{$^{2}$WeChat AI, Tencent Inc.}
}

\begin{document}

\maketitle

\lhead{Preprint}
\blfootnote{$*$~Equal contribution. Work done at WeChat AI.}
\blfootnote{$^{\dagger}$~Corresponding authors: \textless{\href{mailto:zhangwzh@xjtu.edu.cn}{ zhangwzh@xjtu.edu.cn}},~{\href{mailto:withtomzhou@tencent.com}{withtomzhou@tencent.com}}\textgreater}
\blfootnote{$^{\ddagger}$~Project lead: Liyan Xu~\textless{\href{mailto:liyanlxu@tencent.com}{liyanxu.nlp@gmail.com}}\textgreater}

\begin{abstract}
Large reasoning models achieve strong performance on challenging tasks, but their long chain-of-thought often incurs substantial inference costs by continuing to deliberate after sufficient information has been obtained. Existing reasoning compression methods primarily operate on reasoning trajectories, paying less attention to the local decision of whether further reasoning is necessary given the current state. We propose LoCo, a local preference learning method for reasoning compression. LoCo detects candidate overthinking states from the model's own traces, freezes the corresponding reasoning prefixes, regenerates shorter verified continuations from the same states, and trains on the resulting state-conditioned preference pairs with DPO. This formulation teaches the model when to continue reasoning and when to conclude, rather than simply encouraging shorter trajectories overall. We construct training data entirely from self-generated mathematical traces and evaluate LoCo on five mathematical and three code generation benchmarks. LoCo reduces completion length by 41.7\%--45.4\% on mathematics and 23.4\%--36.3\% on code, while maintaining or improving accuracy and achieving the best AUC across all code benchmarks. LoCo consistently improves the accuracy-efficiency trade-off on mathematics and transfers effectively to code generation despite math-only training, demonstrating the cross-domain generality of local preference learning for reasoning compression.
The code and model will be publicly released at \url{https://github.com/MuyaoYuan/LoCo}.
\end{abstract}

\section{Introduction}
Large reasoning models (LRMs)~\citep{guo2025deepseek} have made remarkable progress on complex reasoning tasks such as mathematical problem solving and algorithmic programming. This progress is largely driven by extended chain-of-thought (CoT) reasoning, in which models generate thousands of tokens for hypothesis formation, calculation, revision, and verification before arriving at an answer~\citep{muennighoff2025s1}. Such extended reasoning comes at a substantial inference cost, significantly increasing decoding latency and memory consumption during deployment. More importantly, longer reasoning does not necessarily lead to higher accuracy, as models may continue reasoning beyond what is necessary to reach a correct answer~\citep{chen2024not,sui2025stop}. This tendency to engage in excessive reasoning is commonly referred to as \textit{overthinking}.

Recent studies indicate that when the task is solvable by the model, overthinking is more closely related to model preference than reasoning capability~\citep{zhang2026llms,peng2025revisiting}. A model may get the correct answer early in the reasoning process yet continue generating unnecessary reasoning, reflecting a preference to keep thinking rather than to stop when sufficient reasoning has been reached. The problem therefore lies in when the model chooses to stop, rather than whether it can reach the correct answer~\citep{chen2024not,zhang2026llms}. Mitigating overthinking thus calls for aligning models toward concise reasoning behaviors, which can be learned from carefully curated self-generated reasoning traces~\citep{shen2025dast,zhang2026tokensqueeze}.

Toward this objective, existing self-generated methods derive compression supervision in two main ways, through trajectory-level signals and fine-grained operations within the reasoning process.
Trajectory-level signals are derived from the reasoning trajectory as a whole, using global properties of complete solutions to establish a preference for more concise reasoning. Specifically, one line of work~\citep{team2025kimi,luo2026o1,shen2025dast,ma2025cot} constructs preference pairs by comparing shorter and longer correct solutions, encouraging the model to favor the shorter one, while another~\citep{arora2026training, aggarwal2025l1, hu2026smartthinker} incorporates length preferences into reinforcement learning by rewarding shorter correct reasoning. 

Fine-grained operations provide a more localized alternative by modifying parts of an existing reasoning trajectory. Specifically, these methods remove redundant content~\citep{xia2025tokenskip}, compress individual reasoning steps~\citep{jang2024verbosity}, or shorten continuations from intermediate reasoning states to construct training data for more concise reasoning~\citep{lai2024step,jin2025recut}. Such operations are well aligned with recent findings~\citep{xu2026far} that LLM reasoning proceeds through local and incremental state transitions rather than a globally planned process. However, existing fine-grained methods still treat the reasoning trajectory itself as the object of compression. They focus on which parts of the trajectory should be removed or shortened, rather than whether further reasoning is necessary given the model's current state. This distinction is crucial because the appropriate compression decision can vary across reasoning states: the model may need to continue reasoning at one state but should conclude at another.

Building on this observation, we propose \textbf{LoCo} (\textbf{Lo}cal \textbf{Co}mpression), a local preference learning method for reasoning compression. LoCo identifies overthinking states from the model's own reasoning process and constructs state-conditioned preferences by contrasting alternative continuations from the same reasoning state. The resulting supervision is grounded in the model's own reasoning trajectories and actual reasoning behavior. By making the compression decision conditional on the current reasoning state, LoCo encourages the model to conclude when further reasoning is unnecessary while preserving its ability to continue reasoning when additional computation is needed. In this way, LoCo avoids indiscriminate compression across the reasoning trajectory and reduces the risk of compromising the model's reasoning capability on more challenging problems.

We construct training data from self-generated traces and evaluate LoCo on five mathematical reasoning benchmarks and three coding benchmarks spanning a broad range of difficulty. The experiments show that LoCo consistently achieves the strongest efficiency--performance trade-off on mathematics, with especially clear advantages on the harder datasets where both high compression and high accuracy are required. The same trained model also transfers effectively to coding benchmarks of different difficulty, supporting the view that LoCo improves chain-of-thought compression by reshaping the model's reasoning preference rather than by injecting additional task-specific knowledge. 
To summarize, our main contributions are threefold:
\begin{itemize}
\item We show that effective reasoning compression requires state-aware decisions about when further reasoning is unnecessary, rather than simply favoring shorter trajectories.
\item We introduce LoCo, a local preference learning framework that identifies overthinking states and learns continue-or-conclude preferences from verified continuations sharing the same reasoning prefix.
\item We demonstrate the effectiveness of LoCo across mathematical reasoning and code generation, with consistent improvements in the efficiency-performance trade-off and strong cross-domain generalization.
\end{itemize}

\section{Related Works}

\subsection{Trajectory-Level Compression}

Several reasoning compression methods formulate compression supervision at the trajectory level, treating complete reasoning traces as the unit of comparison. Kimi k1.5~\citep{team2025kimi} introduces a Long2Short stage that contrasts correct short and long solutions generated by the model itself and optimizes their relative preference with DPO. O1-Pruner~\citep{luo2026o1} follows a similar paradigm by applying preference fine-tuning to pre-sampled responses, encouraging the model to favor more concise reasoning. CoT-Valve~\citep{ma2025cot} identifies a parameter-space direction for controlling reasoning length, then learns from globally shortened reasoning traces. Another group of methods~\citep{arora2026training, aggarwal2025l1, hu2026smartthinker} incorporates reasoning length into the reinforcement learning objective, jointly rewarding solution quality and efficient token usage under a prescribed length budget. 
Collectively, these studies show that reasoning can often be shortened without substantially affecting problem-solving performance. However, these methods generally compare or optimize complete reasoning trajectories, which provides an overall preference for shorter solutions but offers less information about where the unnecessary reasoning occurs. Two solutions may reach the same answer through different reasoning paths, making it difficult to determine which part of the reasoning can be safely shortened from their overall length or preference alone. This motivates a more fine-grained approach that identifies unnecessary deliberation at specific points in the reasoning process.

\subsection{Fine-Grained Compression}

Fine-grained operations offer a more localized way to reduce reasoning by modifying selected parts of an existing trajectory. These methods remove redundant content, compress individual reasoning steps, or shorten continuations from intermediate states, in line with recent findings that LLM reasoning proceeds through local and incremental state transitions rather than a globally planned process~\citep{xu2026far}. One line of work removes redundant content from the trajectory itself. TokenSkip~\citep{xia2025tokenskip} prunes verbose tokens, Verbosity-Aware Rationale Reduction~\citep{jang2024verbosity} removes redundant sentences based on principled likelihood criteria, and Reasoning Shaping~\citep{song2025mitigating} reweights computation-heavy segments during RLVR training. TokenSqueeze~\citep{zhang2026tokensqueeze} combines depth-matched data selection with distribution-aligned token-level linguistic refinement and achieves strong performance on standard benchmarks. Another line operates at the step level. Step-DPO~\citep{lai2024step} introduces preference pairs of correct and erroneous continuations sharing the same prefix, while SSPO~\citep{xu2025sspo} and ReCUT~\citep{jin2025recut} adapt this formulation to reasoning compression and encourage more concise step-level reasoning.
Despite their finer granularity, these approaches mainly derive compression signals by contrasting or modifying existing reasoning content. Such supervision is effective for identifying unnecessary tokens, sentences, or steps, but provides less direct guidance on the decision to continue reasoning from a particular state. LoCo addresses this aspect by anchoring the comparison at a fixed reasoning state and learning preferences over its alternative continuations. This formulation captures the local decision of whether further deliberation is warranted while retaining the original state as a shared context.

\section{Preliminaries}
\label{sec:preliminaries}

\paragraph{Large Reasoning Models and Reasoning States.}
Large reasoning models (LRMs), such as DeepSeek-R1~\citep{guo2025deepseek}, are trained to perform extended reasoning before producing a final answer. Given a problem $q$, a model first generates a chain of thought (CoT)
$S=(s_0,s_1,\dots,s_n)$,
consisting of intermediate hypotheses, calculations, revisions, and verifications, before arriving at the final answer. The reasoning process is typically enclosed by dedicated thinking tags, such as \texttt{<think>} and \texttt{</think>} in DeepSeek-R1~\citep{guo2025deepseek}, with paragraph breaks naturally separating different reasoning moves. We therefore segment the CoT into reasoning steps at paragraph boundaries and denote each step by $s_i$. A reasoning step may correspond to establishing a subgoal, performing a computation, verifying an intermediate result, or drawing a conclusion.

We denote the input prompt associated with question $q$ as $\mathrm{prompt}(q)$, which includes the task instruction and the question itself. For notational convenience, we define the reasoning state after step $s_i$ as
\begin{equation}
P_i = \mathrm{prompt}(q) \,\|\, s_0 \,\|\, \cdots \,\|\, s_i,
\end{equation}
where $\|$ denotes concatenation. Thus, $P_i$ contains all information available after completing $s_i$ and before generating $s_{i+1}$. Each state therefore defines a local decision point at which the model can either continue reasoning or conclude. We view overthinking as continuing to generate additional reasoning after the current state already contains sufficient information to support the answer. This state-based formulation provides the notation used throughout the method.

\section{Method}
\label{sec:method}

\subsection{Overview}
\begin{figure}[t]
  \centering
  \includegraphics[width=0.9\linewidth]{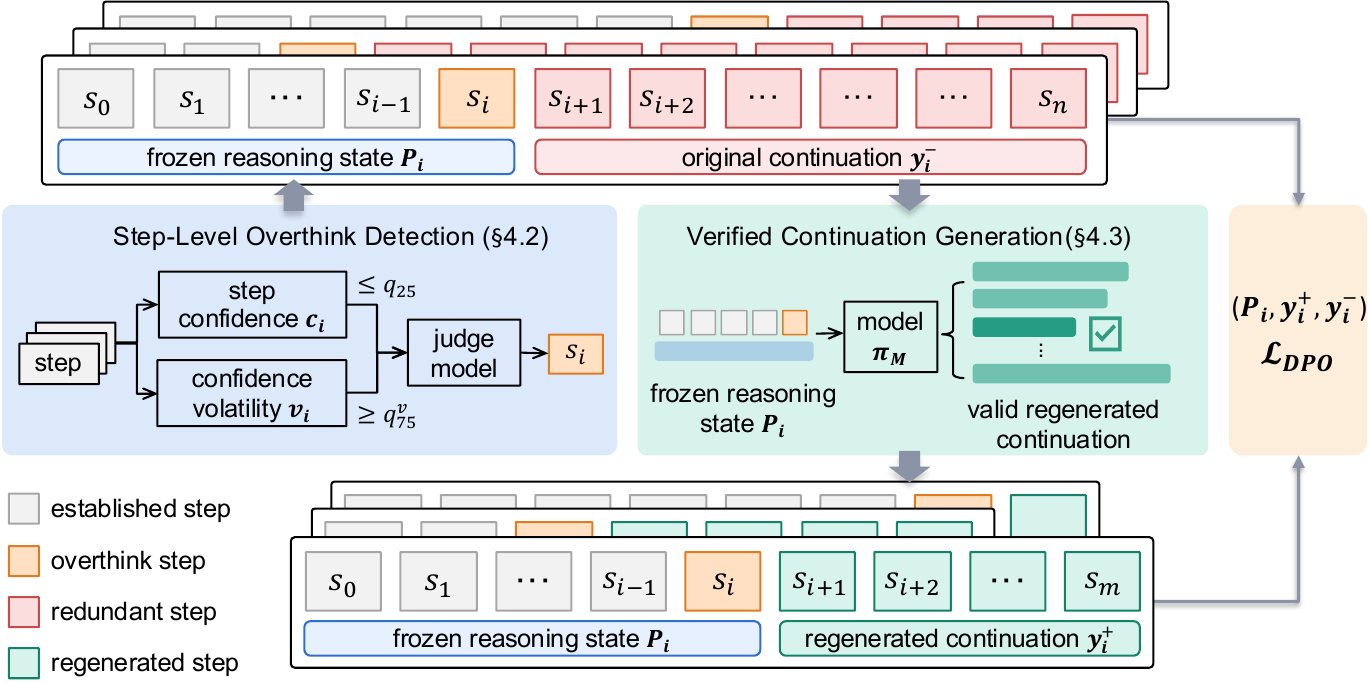}
  \caption{Overview of the LoCo pipeline. Starting from a self-generated reasoning trace, LoCo detects local overthinking states, regenerates shorter correct continuations from those states, and converts them into local preference pairs for training.}
  \label{fig:overview}
\end{figure}

LoCo converts a reasoning model's own overthinking traces into state-anchored preference supervision through three stages, as illustrated in Figure~\ref{fig:overview}. First, local overthink detection screens candidate reasoning states using a two-stage filter in Section~\ref{sec:local_overthink_detection}. Second, verified continuation generation samples shorter correct continuations from each detected state in Section~\ref{sec:prefix_frozen_regeneration}. Third, local preference learning contrasts the regenerated continuation with the original continuation from the same state for DPO training in Section~\ref{sec:counterfactual_preference}. Using the notation introduced in Section~\ref{sec:preliminaries}, this process produces a set of triples $\{(P_i, y_i^{+}, y_i^{-})\}$, where $P_i$ denotes the frozen reasoning state after step $s_i$, $y_i^{+}$ denotes a shorter correct continuation regenerated from $P_i$, and $y_i^{-}$ denotes the original continuation that follows $P_i$. We denote the reasoning model being trained as $M$.

\subsection{Step-Level Overthink Detection}
\label{sec:local_overthink_detection}

We identify reasoning states after which the model begins to overthink using its predictive confidence. Since semantic heuristics can be unreliable and step-level annotation is costly, we first screen candidate states with model confidence and then verify them with a stronger judge model.

\paragraph{Confidence Screening.}
We feed the complete chain of thought back to $M$ and score it under its own next-token distribution. Let $S$ denote the CoT token sequence, $S_{<\tau}$ its prefix preceding position $\tau$, and $\pi_M(\cdot\mid S_{<\tau})$ the corresponding next-token distribution. For each position, we record
\begin{equation}
p_\tau^{\max}=\max_{v\in\mathcal{V}}\pi_M(v\mid S_{<\tau}),
\end{equation}
and let $\mathcal{T}_i$ denote the token positions in step $s_i$, with steps delimited by paragraph breaks. The step confidence is
\begin{equation}
c_i=\frac{1}{|\mathcal{T}_i|}\sum_{\tau\in\mathcal{T}_i}p_\tau^{\max}.
\end{equation}
We measure confidence volatility over a sliding window of recent steps:
\begin{equation}
v_i=\frac{1}{|W_i|}\sum_{j\in W_i}(c_j-\bar{c}_i)^2,
\end{equation}
where $W_i$ contains the most recent steps ending at $i$. We use the minimal window $|W_i|=2$, giving
\begin{equation}
v_i=\frac{(c_i-c_{i-1})^2}{4},
\end{equation}
with $v_0=0$. Thus, $c_i$ measures the model's average predictive confidence within step $s_i$, while $v_i$ captures its change across consecutive steps. Averaging over tokens makes the signal less sensitive to isolated low-probability tokens caused by punctuation or formatting.

\paragraph{Overthink Detection.}
We normalize both signals using the model's own behavior. Let $q_{25}$ and $q_{75}$ be the 25th and 75th percentiles of step confidence, and $q^v_{25}$ and $q^v_{75}$ the corresponding percentiles of volatility, computed once from CoT traces self-generated by $M$. Following the regime analysis of \citet{li2026efficient}, we identify overthinking states as those whose most recent completed step has simultaneously low confidence and high volatility:
\begin{equation}
\mathcal{O}
=
\left\{
i:\ c_i\le q_{25}\ \wedge\ v_i\ge q^v_{75}
\right\}.
\label{eq:mode}
\end{equation}

The conjunction of the two signals distinguishes overthinking from ordinary uncertainty or state transitions. Low confidence can arise from a genuinely difficult sub-problem, while high volatility can accompany a legitimate change in reasoning. Their combination instead identifies states where the model is both uncertain and unstable.
The detector is causal by construction: both $c_i$ and $v_i$ are computed from information available by the end of step $s_i$, whereas truncation begins only from $s_{i+1}$. Hence, the detector never uses the continuation it removes.

\paragraph{Judge Re-checking.}
Confidence alone cannot reliably distinguish redundant deliberation from genuine difficulty. We therefore re-examine each flagged state with a stronger judge model $J$. The two stages use different evidence. The screen identifies candidates solely from the confidence statistics $(c_i,v_i)$, while the judge reads the current reasoning state and determines whether the next step is redundant. The next step is classified as \textsc{overthink} if it re-deliberates on information that has already been established, \textsc{normal} if it makes substantive progress, or \textsc{final} if it presents the answer. This distinction matters because truncation from $P_i$ should remove only redundant continuation. Removing a \textsc{normal} step discards useful progress, while removing a \textsc{final} step may discard the answer itself. This case is particularly relevant because reasoning models often conclude their CoT by restating or boxing the final answer, which may appear repetitive but is necessary for answer extraction.

For each $i\in\mathcal{O}$, the judge receives the problem $q$, the frozen state $P_i$, the next step $s_{i+1}$, and the screen prediction $\hat{\ell}_i$. It returns a corrected role and a rationale:
\begin{equation}
\ell_i
=
J\bigl(q,\;P_i,\;s_{i+1},\;\hat{\ell}_i\bigr)
\in\mathcal{L},
\qquad
\mathcal{L}
=
\bigl\{
\textsc{overthink},
\textsc{normal},
\textsc{final}
\bigr\},
\end{equation}
where $\hat{\ell}_i=\textsc{overthink}$ for all screened candidates. We retain only the states whose next step is confirmed as redundant by the judge:
\begin{equation}
\mathcal{O}
\leftarrow
\bigl\{
i\in\mathcal{O}:\ell_i=\textsc{overthink}
\bigr\}.
\end{equation}

\subsection{Verified Continuation Generation}
\label{sec:prefix_frozen_regeneration}

For each $i\in\mathcal{O}$, we freeze the reasoning state $P_i$ and sample $k$ independent continuations from $M$ conditioned on that state:
\begin{equation}
y_i^{(1)},\ldots,y_i^{(k)}
\sim
\pi_M\bigl(\,\cdot\mid P_i\,\bigr),
\label{eq:regen}
\end{equation}
A continuation is considered valid if it completes the reasoning template and produces an answer symbolically equivalent to the gold answer $z$, denoted by $y\models z$. Among all valid continuations, we select the shortest one:
\begin{equation}
\mathcal{V}_i
=
\bigl\{
j:y_i^{(j)}\models z
\bigr\},
\qquad
y_i^{+}
=
\operatorname*{arg\,min}_{j\in\mathcal{V}_i}
\bigl|y_i^{(j)}\bigr|,
\label{eq:select}
\end{equation}
where $|\cdot|$ denotes the number of tokens. If $\mathcal{V}_i$ is empty, we discard the detection point. Otherwise, we define the original continuation from the same state as $y_i^{-}=(s_{i+1},s_{i+2},\ldots)$ and retain the point only when the selected continuation is strictly shorter:
\begin{equation}
\mathcal{O}
\leftarrow
\bigl\{
i\in\mathcal{O}:
|y_i^{+}|<|y_i^{-}|
\bigr\}.
\label{eq:regen_refine}
\end{equation}

\subsection{Local Preference Learning}
\label{sec:counterfactual_preference}

For each surviving detection point $i\in\mathcal{O}$, we construct the preference triple $(P_i,\;y_i^{+},\;y_i^{-})$: the two continuations complete the identical frozen state $P_i$, the chosen $y_i^{+}$ is the shortest verified continuation, and the rejected $y_i^{-}$ is the original continuation. Writing $g_i=|y_i^{-}|-|y_i^{+}|$ for the token savings of the pair, we retain a pair only when the savings reach a minimum $\delta$, the prompt fits within $c_p$ tokens, and each continuation fits within $c_r$ tokens. Since larger savings indicate clearer compression signals, at most $K$ pairs are kept per question, preventing questions with many detection points from dominating the training data:
\begin{equation}
\mathcal{D}=\bigl\{\,\bigl(P_i,\;y_i^{+},\;y_i^{-}\bigr)\::\:i\in\mathcal{S}\,\bigr\},
\qquad
\mathcal{S}=\bigl\{\,i\in\mathcal{O}:\;g_i\ge\delta,\;\bigl|P_i\bigr|\le c_p,\;\bigl|y_i^{\pm}\bigr|\le c_r\,\bigr\},
\label{eq:dset}
\end{equation}
where $\mathcal{S}$ is truncated within each question to its $K$ largest-$g_i$ elements.

The shared prefix provides a natural state-conditional preference objective. With $x=P_i$, we optimize the DPO objective
\begin{equation}
\mathcal{L}_{\mathrm{DPO}}
=
-\mathbb{E}_{(x,\,y^{+},\,y^{-})\sim\mathcal{D}}
\left[
\log \sigma
\left(
\beta
\left[
\log \frac{\pi_\theta(y^{+}|x)}{\pi_{\mathrm{ref}}(y^{+}|x)}
-
\log \frac{\pi_\theta(y^{-}|x)}{\pi_{\mathrm{ref}}(y^{-}|x)}
\right]
\right)
\right],
\label{eq:dpo}
\end{equation}
where $\pi_{\mathrm{ref}}$ is the frozen reference policy, and each log-likelihood is computed as the sum of token log-probabilities over the corresponding continuation, $\log\pi(y|x)=\sum_{t}\log\pi\bigl(y_t\,\big|\,x,\,y_{<t}\bigr)$. Since both branches share the same prefix $P_i$, the preference comparison is conditioned on the identical reasoning state and directly supervises which continuation the model should prefer after reaching that state. In this way, LoCo learns a local continue-or-conclude preference: once the model has reached state $P_i$, it learns when to conclude with a shorter valid continuation instead of continuing with the original redundant reasoning.

\section{Experiments}
\label{sec:experiments}

\subsection{Experimental Setup}

\paragraph{Data Construction and Training Configurations.}
The preference data are constructed from self-generated traces of DeepSeek-R1-Distill-Qwen-7B on the math-only training set used by ~\citet{zhang2026tokensqueeze}. For each query, we sample three rollouts with temperature 0.6 and top-$p$ 0.95, and use the shortest correct rollout as the original trajectory. We then re-examine each flagged reasoning step with Qwen3-30B as the judge. For each detected step, we regenerate $k=8$ continuations using the same temperature and top-$p$, and retain preference pairs with at least $\delta=300$ tokens of savings, while limiting each question to at most $K=3$ pairs. The prompt and continuation lengths are capped at $c_p=6$k and $c_r=12$k tokens, respectively.

We train DeepSeek-R1-Distill-Qwen-7B with LoCo for two epochs using a LoRA adapter. The adapter has rank 256, scaling factor $\alpha=512$, and dropout 0.05 on the q/k/v/o attention projections, while the base model remains frozen. We use an effective batch size of 64 and a learning rate of $1\times10^{-5}$, with a cosine schedule and a warmup ratio of 0.1. Optimization uses AdamW with a weight decay of 0.01 and gradient clipping of 1.0. The DPO temperature is set to $\beta=0.1$, with the base model serving as the reference policy. Training is conducted on four H20 GPUs with a maximum sequence length of 14,336 tokens.

\paragraph{Evaluation.}
We evaluate LoCo on mathematical reasoning and code generation. For mathematical reasoning, we use five benchmarks spanning different difficulty levels: MATH-500~\citep{lightman2024let} with 500 problems, AMC 2023 with 40 problems, Olympiad~\citep{he2024olympiadbench} with 675 problems, AIME 2024 with 30 problems, and AIME 2025 with 30 problems. For code generation, we use MBPP~\citep{austin2021program} sanitized with 257 problems, LiveCodeBench~\citep{jain2025livecodebench} release v5 with 167 problems, and CodeContests~\citep{li2022competition} with 165 problems. We sample eight completions per problem for all benchmarks.

Evaluation is performed with vLLM using the LoRA adapter and batched generation, with temperature 0.6 and top-$p$ 0.95. We allow up to 32,768 newly generated tokens with a context length of 36,864, and generate until EOS without truncating at \texttt{</think>}. 
For mathematical reasoning, we extract the final answer from the last \texttt{\textbackslash boxed\{\}} expression in the complete model output, and check it for symbolic equivalence using \texttt{math\_verify}. For code generation, we extract the program from the last Python code block and evaluate it by sandboxed execution, following the official protocols for LiveCodeBench and MBPP and the AlphaCode judging protocol for CodeContests with a numerical tolerance of $10^{-5}$.
We report Accuracy, Tokens, and AUC following \citet{zhang2026tokensqueeze}, where AUC denotes the area under the accuracy versus token budget curve obtained by varying the maximum allowed reasoning length. All baselines~\citep{guo2025deepseek,shen2025dast,zhang2026tokensqueeze} are evaluated from their released full weights without
modification.

\subsection{Evaluation on Mathematical Reasoning Benchmarks}

\newcommand{\stdv}[1]{{\fontsize{6.4pt}{7.0pt}\selectfont$\pm$#1}}

We first evaluate LoCo on mathematical reasoning benchmarks that span a wide difficulty range, from MATH-500 and AMC23 to Olympiad, AIME24, and AIME25. This setting tests whether the method can compress reasoning aggressively without sacrificing performance as problem difficulty increases.
\begin{table}[t]
  \centering
  \caption{Mathematical reasoning results across five benchmarks of increasing difficulty. The table reports absolute performance and base-relative changes under a shared evaluation protocol. LoCo achieves the best overall trade-off, combining the strongest compression with consistently comparable or improved accuracy across difficulty levels.}
  \label{tab:main_results}
  \setlength{\tabcolsep}{3.0pt}
  \renewcommand{\arraystretch}{1.08}
  \fontsize{7.5pt}{9.6pt}\selectfont
  \begin{tabular}{@{}l|ccc|ccc|ccc@{}}
    \toprule
    \rowcolor{TableHeader}
    \multirow{2}{*}{\textbf{Method}} & \multicolumn{3}{c}{\textbf{MATH-500}} & \multicolumn{3}{c}{\textbf{AMC23}} & \multicolumn{3}{c}{\textbf{Olympiad}} \\
    \cmidrule(lr){2-4}\cmidrule(lr){5-7}\cmidrule(lr){8-10}
    \rowcolor{TableHeader}
    & \textbf{Acc (\%)} & \textbf{Token} & \textbf{AUC (\%)} & \textbf{Acc (\%)} & \textbf{Token} & \textbf{AUC (\%)} & \textbf{Acc (\%)} & \textbf{Token} & \textbf{AUC (\%)} \\
    \midrule
    Base & \textbf{91.88}\stdv{1.20} & 4015\stdv{177} & 82.64\stdv{1.12} & 90.31\stdv{2.48} & 6285\stdv{516} & 76.95\stdv{2.07} & 61.33\stdv{0.41} & 8972\stdv{130} & 51.28\stdv{0.37} \\
    \addlinespace[2pt]
    \multirow{2}{*}{DAST} & 91.38\stdv{0.79} & 3720\stdv{119} & 83.95\stdv{0.66} & 88.12\stdv{4.17} & 5981\stdv{304} & 76.89\stdv{3.22} & 59.91\stdv{0.73} & 9155\stdv{313} & 51.86\stdv{0.59} \\
    & {\scriptsize(-0.50)} & {\scriptsize($\downarrow$7.4\%)} & {\scriptsize(+1.31)} & {\scriptsize(-2.19)} & {\scriptsize($\downarrow$4.8\%)} & {\scriptsize(-0.06)} & {\scriptsize(-1.42)} & {\scriptsize($\uparrow$2.0\%)} & {\scriptsize(+0.58)} \\
    \addlinespace[3pt]
    \multirow{2}{*}{TokenSqueeze} & 91.18\stdv{0.89} & 2442\stdv{52} & 85.92\stdv{0.73} & 88.44\stdv{4.62} & 4350\stdv{582} & 79.42\stdv{4.01} & 60.98\stdv{0.95} & 6149\stdv{105} & 54.29\stdv{0.78} \\
    & {\scriptsize(-0.70)} & {\scriptsize($\downarrow$39.2\%)} & {\scriptsize(+3.28)} & {\scriptsize(-1.87)} & {\scriptsize($\downarrow$30.8\%)} & {\scriptsize(+2.47)} & {\scriptsize(-0.35)} & {\scriptsize($\downarrow$31.5\%)} & {\scriptsize(+3.01)} \\
    \addlinespace[3pt]
    \ourlabel & \ourcell{91.80\stdv{0.88}} & \ourcell{\textbf{2243}\stdv{83}} & \ourcell{\textbf{86.56}\stdv{0.76}} & \ourcell{\textbf{91.56}\stdv{2.65}} & \ourcell{\textbf{3663}\stdv{270}} & \ourcell{\textbf{83.44}\stdv{2.12}} & \ourcell{\textbf{62.78}\stdv{0.64}} & \ourcell{\textbf{4968}\stdv{133}} & \ourcell{\textbf{56.61}\stdv{0.61}} \\
    \ourblank & \ourcell{{\scriptsize(-0.08)}} & \ourcell{{\scriptsize\textbf{($\downarrow$44.1\%)}}} & \ourcell{{\scriptsize\textbf{(+3.92)}}} & \ourcell{{\scriptsize\textbf{(+1.25)}}} & \ourcell{{\scriptsize\textbf{($\downarrow$41.7\%)}}} & \ourcell{{\scriptsize\textbf{(+6.49)}}} & \ourcell{{\scriptsize\textbf{(+1.45)}}} & \ourcell{{\scriptsize\textbf{($\downarrow$44.6\%)}}} & \ourcell{{\scriptsize\textbf{(+5.33)}}} \\
    \bottomrule
  \end{tabular}

  \vspace{0.65em}

  \begin{tabular}{@{}l|ccc|ccc@{}}
    \toprule
    \rowcolor{TableHeader}
    \multirow{2}{*}{\textbf{Method}} & \multicolumn{3}{c}{\textbf{AIME24}} & \multicolumn{3}{c}{\textbf{AIME25}} \\
    \cmidrule(lr){2-4}\cmidrule(lr){5-7}
    \rowcolor{TableHeader}
    & \textbf{Acc (\%)} & \textbf{Token} & \textbf{AUC (\%)} & \textbf{Acc (\%)} & \textbf{Token} & \textbf{AUC (\%)} \\
    \midrule
    Base & 50.83\stdv{6.61} & 13493\stdv{1367} & 40.33\stdv{5.00} & 37.92\stdv{5.62} & 15134\stdv{920} & 30.29\stdv{3.82} \\
    \addlinespace[2pt]
    \multirow{2}{*}{DAST} & 50.83\stdv{6.36} & 14423\stdv{2075} & 41.40\stdv{4.81} & 37.92\stdv{4.34} & 15652\stdv{1637} & 31.66\stdv{3.79} \\
    & {\scriptsize($\pm$0.00)} & {\scriptsize($\uparrow$6.9\%)} & {\scriptsize(+1.07)} & {\scriptsize($\pm$0.00)} & {\scriptsize($\uparrow$3.4\%)} & {\scriptsize(+1.37)} \\
    \addlinespace[3pt]
    \multirow{2}{*}{TokenSqueeze} & 48.75\stdv{4.69} & 10457\stdv{831} & 40.14\stdv{4.05} & 35.83\stdv{4.96} & 11055\stdv{585} & 31.20\stdv{4.12} \\
    & {\scriptsize(-2.08)} & {\scriptsize($\downarrow$22.5\%)} & {\scriptsize(-0.19)} & {\scriptsize(-2.09)} & {\scriptsize($\downarrow$27.0\%)} & {\scriptsize(+0.91)} \\
    \addlinespace[3pt]
    \ourlabel & \ourcell{\textbf{54.17}\stdv{4.27}} & \ourcell{\textbf{7364}\stdv{399}} & \ourcell{\textbf{46.82}\stdv{3.89}} & \ourcell{\textbf{40.42}\stdv{3.75}} & \ourcell{\textbf{8291}\stdv{250}} & \ourcell{\textbf{34.55}\stdv{3.13}} \\
    \ourblank & \ourcell{{\scriptsize\textbf{(+3.34)}}} & \ourcell{{\scriptsize\textbf{($\downarrow$45.4\%)}}} & \ourcell{{\scriptsize\textbf{(+6.49)}}} & \ourcell{{\scriptsize\textbf{(+2.50)}}} & \ourcell{{\scriptsize\textbf{($\downarrow$45.2\%)}}} & \ourcell{{\scriptsize\textbf{(+4.26)}}} \\
    \bottomrule
  \end{tabular}
\end{table}

Table~\ref{tab:main_results} shows a consistent pattern across difficulty levels. On the easier benchmarks, LoCo already matches or improves upon the base model while removing 41.7\%--44.1\% of the generated tokens. More importantly, this advantage becomes even clearer on the harder benchmarks: on Olympiad, AIME24, and AIME25, LoCo simultaneously attains the highest accuracy, the highest AUC, and a 44.6\%--45.4\% reduction in completion length. This behavior distinguishes LoCo from both baselines. DAST provides only mild compression and can even increase completion length on harder benchmarks, while TokenSqueeze compresses aggressively but loses more accuracy as difficulty rises. Overall, the results indicate that LoCo preserves reasoning quality under substantial compression and remains particularly strong exactly where preserving difficult reasoning matters most.

\subsection{Evaluation on Coding Benchmarks}

We next evaluate LoCo on code generation to test whether the learned compression behavior transfers beyond the math-only training queries described above. This provides a strict generalization setting. If the method still improves performance across coding benchmarks of different difficulty, the gain is more plausibly explained by a change in the model's reasoning preference, specifically when to continue and when to conclude, rather than by injecting additional domain knowledge.
\begin{table}[t]
  \centering
  \caption{Code generation results on three benchmarks of increasing difficulty, evaluated after training on a math-only query pool. The table reports absolute performance and base-relative changes under a shared evaluation protocol. LoCo retains the strongest overall advantage, showing that the learned compression preference generalizes beyond the training domain.}
  \label{tab:coding_results}
  \setlength{\tabcolsep}{2.2pt}
  \renewcommand{\arraystretch}{1.08}
  \fontsize{7.5pt}{9.6pt}\selectfont
  \begin{tabular}{@{}l|ccc|ccc|ccc@{}}
    \toprule
    \rowcolor{TableHeader}
    \multirow{2}{*}{\textbf{Method}} & \multicolumn{3}{c}{\textbf{MBPP}} & \multicolumn{3}{c}{\textbf{LiveCodeBench}} & \multicolumn{3}{c}{\textbf{CodeContests}} \\
    \cmidrule(lr){2-4}\cmidrule(lr){5-7}\cmidrule(lr){8-10}
    \rowcolor{TableHeader}
    & \textbf{Acc (\%)} & \textbf{Token} & \textbf{AUC (\%)} & \textbf{Acc (\%)} & \textbf{Token} & \textbf{AUC (\%)} & \textbf{Acc (\%)} & \textbf{Token} & \textbf{AUC (\%)} \\
    \midrule
    Base & \textbf{84.05}\stdv{0.91} & 3403\stdv{234} & 78.42\stdv{0.83} & 41.77\stdv{1.10} & 12319\stdv{352} & 32.86\stdv{0.87} & 18.48\stdv{2.17} & 14900\stdv{223} & 13.87\stdv{1.50} \\
    \addlinespace[2pt]
    \multirow{2}{*}{DAST} & 81.13\stdv{1.87} & 4097\stdv{444} & 76.32\stdv{1.68} & 39.82\stdv{2.50} & 14158\stdv{476} & 31.88\stdv{1.66} & 16.06\stdv{2.42} & 17757\stdv{623} & 12.61\stdv{1.79} \\
    & {\scriptsize(-2.92)} & {\scriptsize($\uparrow$20.4\%)} & {\scriptsize(-2.10)} & {\scriptsize(-1.95)} & {\scriptsize($\uparrow$14.9\%)} & {\scriptsize(-0.98)} & {\scriptsize(-2.42)} & {\scriptsize($\uparrow$19.2\%)} & {\scriptsize(-1.26)} \\
    \addlinespace[3pt]
    \multirow{2}{*}{TokenSqueeze} & 81.37\stdv{2.18} & 2196\stdv{131} & 77.40\stdv{1.90} & 41.69\stdv{2.22} & 9670\stdv{200} & 35.03\stdv{1.38} & 19.09\stdv{0.97} & 11677\stdv{305} & 15.54\stdv{0.81} \\
    & {\scriptsize(-2.68)} & {\scriptsize($\downarrow$35.5\%)} & {\scriptsize(-1.02)} & {\scriptsize(-0.08)} & {\scriptsize($\downarrow$21.5\%)} & {\scriptsize(+2.17)} & {\scriptsize(+0.61)} & {\scriptsize($\downarrow$21.6\%)} & {\scriptsize(+1.67)} \\
    \addlinespace[3pt]
    \ourlabel & \ourcell{\textbf{84.05}\stdv{1.94}} & \ourcell{\textbf{2167}\stdv{119}} & \ourcell{\textbf{79.95}\stdv{1.95}} & \ourcell{\textbf{42.51}\stdv{2.81}} & \ourcell{\textbf{9278}\stdv{204}} & \ourcell{\textbf{35.64}\stdv{2.15}} & \ourcell{\textbf{20.61}\stdv{1.07}} & \ourcell{\textbf{11413}\stdv{243}} & \ourcell{\textbf{16.78}\stdv{0.98}} \\
    \ourblank & \ourcell{{\scriptsize($\pm0.00$)}} & \ourcell{{\scriptsize\textbf{($\downarrow$36.3\%)}}} & \ourcell{{\scriptsize\textbf{(+1.53)}}} & \ourcell{{\scriptsize\textbf{(+0.74)}}} & \ourcell{{\scriptsize\textbf{($\downarrow$24.7\%)}}} & \ourcell{{\scriptsize\textbf{(+2.78)}}} & \ourcell{{\scriptsize\textbf{(+2.13)}}} & \ourcell{{\scriptsize\textbf{($\downarrow$23.4\%)}}} & \ourcell{{\scriptsize\textbf{(+2.91)}}} \\
    \bottomrule
  \end{tabular}
\end{table}

Table~\ref{tab:coding_results} shows that this transfer is consistent across coding benchmarks of different difficulty. On MBPP, LoCo matches the base model in accuracy while reducing tokens by 36.3\% and improving AUC by 1.53 points. On the more challenging LiveCodeBench and CodeContests benchmarks, it not only produces substantially shorter completions, but also improves both accuracy and AUC over all baselines. Since the training queries are mathematical rather than coding tasks, these gains support the interpretation that LoCo works by reshaping the model's preference over reasoning length, instead of relying on additional task-specific knowledge injected during training.

\begin{figure}[t]
  \centering
  \begin{minipage}[t]{0.42\linewidth}
    \centering
    \includegraphics[width=\linewidth]{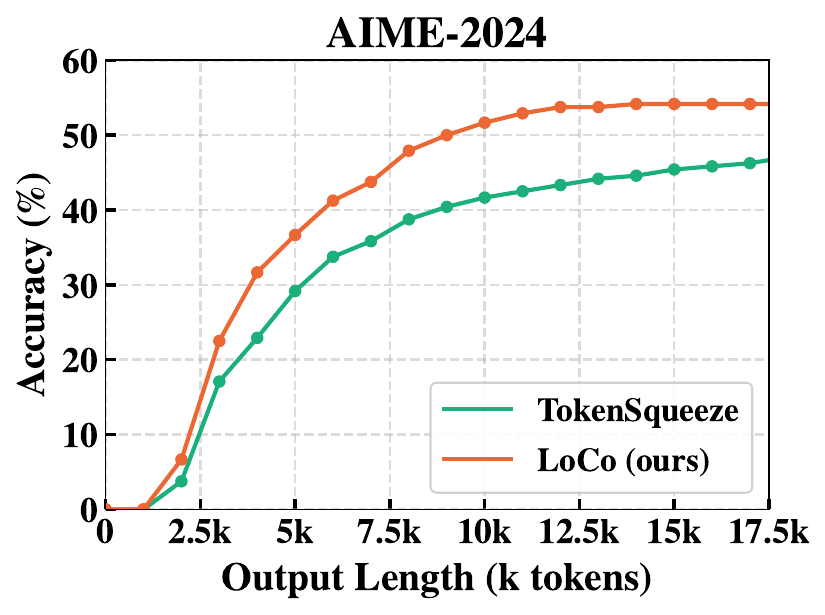}
    {\footnotesize\textbf{(a)} AIME24}
  \end{minipage}
  \hspace{0.02\linewidth}
  \begin{minipage}[t]{0.42\linewidth}
    \centering
    \includegraphics[width=\linewidth]{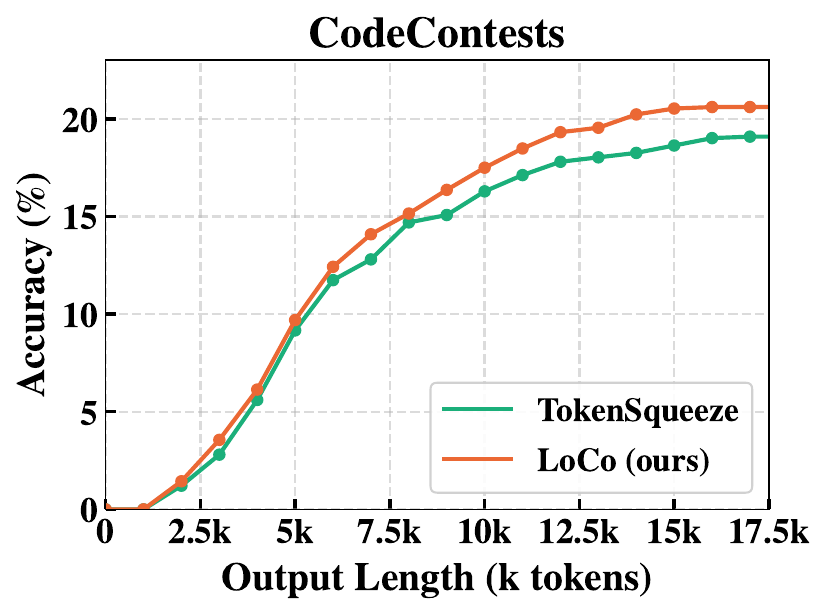}
    {\footnotesize\textbf{(b)} CodeContests}
  \end{minipage}
  \caption{Accuracy versus token budget on AIME24 and CodeContests. LoCo achieves a better accuracy and efficiency trade-off than TokenSqueeze on both benchmarks.}
  \label{fig:main_budget_curves}
\end{figure}

Figure~\ref{fig:main_budget_curves} further illustrates the AUC gains in Tables~\ref{tab:main_results} and~\ref{tab:coding_results}. On both benchmarks, LoCo stays above TokenSqueeze over most token budgets, indicating a better overall accuracy--efficiency frontier rather than an advantage at a single cutoff.

\subsection{Ablation Study}

To understand which components are responsible for LoCo's gains, we keep the training setting fixed and vary one design choice at a time. Table~\ref{tab:ablation_results} compares alternative preference sources and cutpoint strategies against the full method. As in the results table, we report absolute values for non-base variants together with their changes relative to the base model.

\begin{table}[t]
  \centering
  \caption{Ablation results on three representative mathematical reasoning benchmarks under matched training settings. The table compares alternative preference sources and cutpoint choices against the full LoCo design. LoCo yields the strongest overall compression--accuracy trade-off, indicating that both local preference construction and detector-selected cutpoints are important.}
  \label{tab:ablation_results}
  \setlength{\tabcolsep}{2.3pt}
  \renewcommand{\arraystretch}{1.06}
  \fontsize{7.0pt}{8.7pt}\selectfont
  \begin{tabular}{@{}l|ccc|ccc|ccc@{}}
    \toprule
    \rowcolor{TableHeader}
    \multirow{2}{*}{\textbf{Method}} & \multicolumn{3}{c}{\textbf{AMC23}} & \multicolumn{3}{c}{\textbf{AIME24}} & \multicolumn{3}{c}{\textbf{AIME25}} \\
    \cmidrule(lr){2-4}\cmidrule(lr){5-7}\cmidrule(lr){8-10}
    \rowcolor{TableHeader}
    & \textbf{Acc (\%)} & \textbf{Token} & \textbf{AUC (\%)} & \textbf{Acc (\%)} & \textbf{Token} & \textbf{AUC (\%)} & \textbf{Acc (\%)} & \textbf{Token} & \textbf{AUC (\%)} \\
    \midrule
    Base & 90.31\stdv{2.48} & 6285\stdv{516} & 76.95\stdv{2.07} & 50.83\stdv{6.61} & 13493\stdv{1367} & 40.33\stdv{5.00} & 37.92\stdv{5.62} & 15134\stdv{920} & 30.29\stdv{3.82} \\
    \addlinespace[2pt]
    \rowcolor{TableGroup}
    \multicolumn{10}{@{}l}{\textbf{Preference source}} \\
    \multirow{2}{*}{Global} & 90.94\stdv{2.65} & 5957\stdv{562} & 77.66\stdv{2.42} & \textbf{54.17}\stdv{5.56} & 12002\stdv{870} & 42.57\stdv{4.23} & \textbf{40.42}\stdv{5.47} & 12740\stdv{1079} & 32.46\stdv{3.90} \\
    & {\scriptsize(+0.63)} & {\scriptsize($\downarrow$5.2\%)} & {\scriptsize(+0.71)} & {\scriptsize\textbf{(+3.34)}} & {\scriptsize($\downarrow$11.1\%)} & {\scriptsize(+2.24)} & {\scriptsize\textbf{(+2.50)}} & {\scriptsize($\downarrow$15.8\%)} & {\scriptsize(+2.17)} \\
    \addlinespace[3pt]
    \multirow{2}{*}{GPT-5.4 Cont.} & 90.62\stdv{2.59} & 4083\stdv{261} & 81.46\stdv{2.17} & 52.92\stdv{6.53} & 9204\stdv{823} & 44.11\stdv{5.54} & 39.58\stdv{6.53} & 9690\stdv{709} & 33.56\stdv{4.78} \\
    & {\scriptsize(+0.31)} & {\scriptsize($\downarrow$35.0\%)} & {\scriptsize(+4.51)} & {\scriptsize(+2.09)} & {\scriptsize($\downarrow$31.8\%)} & {\scriptsize(+3.78)} & {\scriptsize(+1.66)} & {\scriptsize($\downarrow$36.0\%)} & {\scriptsize(+3.27)} \\
    \addlinespace[2pt]
    \rowcolor{TableGroup}
    \multicolumn{10}{@{}l}{\textbf{Cutpoint design}} \\
    \multirow{2}{*}{RandCut} & 88.44\stdv{2.97} & 3679\stdv{243} & 80.59\stdv{2.40} & 48.33\stdv{5.35} & 7880\stdv{642} & 41.64\stdv{4.57} & 38.75\stdv{5.62} & 9151\stdv{968} & 34.06\stdv{4.39} \\
    & {\scriptsize(-1.87)} & {\scriptsize($\downarrow$41.5\%)} & {\scriptsize(+3.64)} & {\scriptsize(-2.50)} & {\scriptsize($\downarrow$41.6\%)} & {\scriptsize(+1.31)} & {\scriptsize(+0.83)} & {\scriptsize($\downarrow$39.5\%)} & {\scriptsize(+3.77)} \\
    \addlinespace[3pt]
    \multirow{2}{*}{UniCut} & 89.38\stdv{2.59} & 3573\stdv{382} & 81.17\stdv{2.81} & 48.75\stdv{5.02} & 7895\stdv{482} & 42.58\stdv{4.08} & 36.67\stdv{6.17} & 8627\stdv{751} & 31.76\stdv{4.73} \\
    & {\scriptsize(-0.93)} & {\scriptsize($\downarrow$43.2\%)} & {\scriptsize(+4.22)} & {\scriptsize(-2.08)} & {\scriptsize($\downarrow$41.5\%)} & {\scriptsize(+2.25)} & {\scriptsize(-1.25)} & {\scriptsize($\downarrow$43.0\%)} & {\scriptsize(+1.47)} \\
    \addlinespace[3pt]
    \multirow{2}{*}{PosMat} & 89.69\stdv{2.48} & \textbf{3570}\stdv{235} & 82.09\stdv{1.79} & 50.00\stdv{5.63} & 7576\stdv{1056} & 43.66\stdv{4.87} & 35.83\stdv{4.63} & 8683\stdv{720} & 31.50\stdv{3.47} \\
    & {\scriptsize(-0.62)} & {\scriptsize\textbf{($\downarrow$43.2\%)}} & {\scriptsize(+5.14)} & {\scriptsize(-0.83)} & {\scriptsize($\downarrow$43.9\%)} & {\scriptsize(+3.33)} & {\scriptsize(-2.09)} & {\scriptsize($\downarrow$42.6\%)} & {\scriptsize(+1.21)} \\
    \addlinespace[3pt]
    \ourlabel & \ourcell{\textbf{91.56}\stdv{2.65}} & \ourcell{3663\stdv{270}} & \ourcell{\textbf{83.44}\stdv{2.12}} & \ourcell{\textbf{54.17}\stdv{4.27}} & \ourcell{\textbf{7364}\stdv{399}} & \ourcell{\textbf{46.82}\stdv{3.89}} & \ourcell{\textbf{40.42}\stdv{3.75}} & \ourcell{\textbf{8291}\stdv{250}} & \ourcell{\textbf{34.55}\stdv{3.13}} \\
    \ourblank & \ourcell{{\scriptsize\textbf{(+1.25)}}} & \ourcell{{\scriptsize($\downarrow$41.7\%)}} & \ourcell{{\scriptsize\textbf{(+6.49)}}} & \ourcell{{\scriptsize\textbf{(+3.34)}}} & \ourcell{{\scriptsize\textbf{($\downarrow$45.4\%)}}} & \ourcell{{\scriptsize\textbf{(+6.49)}}} & \ourcell{{\scriptsize\textbf{(+2.50)}}} & \ourcell{{\scriptsize\textbf{($\downarrow$45.2\%)}}} & \ourcell{{\scriptsize\textbf{(+4.26)}}} \\
    \bottomrule
  \end{tabular}
\end{table}

\paragraph{Local versus global preferences.}
In the \textit{Global} ablation setting, we keep the same query set, pair-selection thresholds, and training setting as LoCo, but remove the state structure entirely. For each query, we form a trajectory-level preference pair between the shortest and longest correct full rollouts, so the prompt contains only the original query rather than a frozen intermediate state. Compared with this global construction, LoCo concentrates the supervision signal at the specific state where compression becomes appropriate. This difference matters most for compression. Although \textit{Global} reaches similar accuracy on some benchmarks, it reduces tokens by only 5.2\% on AMC23, 11.1\% on AIME24, and 15.8\% on AIME25, far below LoCo's 41.7\%, 45.4\%, and 45.2\%. These results suggest that effective reasoning compression depends not only on preferring shorter reasoning, but also on identifying where compression should occur.

\paragraph{Self-generated versus external continuations.}
In the \textit{GPT-5.4 Cont.} ablation setting, we keep the same cutpoints, frozen prefixes, rejected continuations, pair-selection thresholds, and training setting as LoCo, and change only the chosen continuation. GPT-5.4 regenerates from the same frozen prefix and replaces the response only when it is both correct and shorter. Compared with this external-continuation variant, LoCo performs better overall in compression and AUC, especially on the harder AIME benchmarks. A plausible explanation is distributional alignment. The preference pairs in LoCo are built from continuations generated by the same model being trained, so the supervision remains closer to the model's native reasoning distribution. In contrast, external continuations provide useful signals, but they are less matched to the model's own CoT style and therefore offer a weaker training signal for learning when to stop.

\paragraph{Detector-selected versus heuristic cutpoints.}
In the \textit{RandCut}, \textit{UniCut}, and \textit{PosMat} ablations, we keep the regeneration process, pair-selection thresholds, and training configuration identical to LoCo, and only change how the cutpoints are selected. Each trace is assigned the same number of cutpoints as in LoCo. Specifically, \textit{RandCut} selects cutpoints uniformly at random from the trajectory, while \textit{UniCut} selects them at evenly spaced positions. Moreover, we introduce a position-matched ablation, \textit{PosMat}, which randomly samples cutpoints from the detector's relative-position distribution, isolating the effect of detector-based selection from that of cutpoint location. The shorter continuations are then regenerated and paired following the same procedure as LoCo.
The heuristic variants can still shorten CoT substantially, but they incur larger accuracy losses because they often compress reasoning at positions where the model still needs additional computation. This effect is especially clear on AIME24 and AIME25, where LoCo preserves the strongest accuracy while exceeding the heuristic variants in overall AUC. These comparisons show that LoCo's advantage comes not only from preferring shorter continuations, but from learning that preference at the right reasoning states.

\section{Conclusion}
We introduced LoCo, a local preference learning method for reasoning compression that constructs supervision directly at reasoning states where overthinking occurs. Instead of treating compression as a global preference over complete trajectories, LoCo freezes the current reasoning prefix, regenerates verified shorter continuations from the same state, and learns state-conditioned continue-or-conclude preferences with DPO. Empirically, LoCo shows consistent gains across both mathematical reasoning and code generation. The cross-domain improvements suggest that LoCo primarily reshapes the model's preference over reasoning length rather than adding task-specific knowledge. This indicates that effective reasoning compression requires learning when further reasoning is unnecessary at the appropriate reasoning states, rather than simply favoring shorter trajectories.

\subsection*{AI use statement}
In this work, we used generative AI tools for generating synthetic data sets, cleaning and reformatting datasets, and assisting with translation. We have not used generative AI tools for helping develop theoretical models or conceptual frameworks, proposing or refining hypotheses, designing or providing feedback on research methodology or experiments, implementing methods, supporting qualitative and thematic data analysis, interpreting results, and formulating mathematical claims, providing critical ingredients for proving mathematical claims, assisting in the writing of proofs are not applicable to this work. Additionally, we used generative AI tools for editing the paper to improve readability. We have reviewed all AI-assisted work. Generating and cleaning data using large language models is a primary method of our study design, for which we conducted detailed experiments, and we checked all AI-edited text. We take responsibility for the final content of this work, including text, claims or artifacts produced with the aid of generative AI.

\subsection*{Reproducibility statement}
We provide the methodological and experimental details required to reproduce our results in the main text and appendices. Section~\ref{sec:method} describes the LoCo method, including local overthink detection, verified continuation generation, and local preference learning. Section~\ref{sec:experiments} specifies the training setup, evaluation protocol, benchmark composition, decoding settings, and ablation design used in our experiments. Appendix~\ref{app:implementation} provides additional implementation details, while Sections~\ref{app:evaluation} and~\ref{app:ablation_details} specify the evaluation prompts and the full construction of the ablation variants.

\bibliography{iclr2027_conference}
\bibliographystyle{iclr2027_conference}

\appendix

\section{Implementation Details}
\label{app:implementation}

\subsection{Overthink Detection}
\label{app:detection}

\paragraph{Source traces.}
We generate self-sampled reasoning traces from DeepSeek-R1-Distill-Qwen-7B on the math14k training pool, the math-only training set of \citet{zhang2026tokensqueeze} containing 14{,}027 questions. We sample three rollouts per question at temperature 0.6 and top-$p$ 0.95, and retain the shortest correct rollout as the source trajectory for each question. Detection, judging, and pair construction all operate on this trajectory, so each question contributes at most one trace.

\paragraph{Confidence screening.}
The complete CoT is fed back to the model as a prompt, and we record the top-1 token probability $p_\tau^{\max}$ at every position via batched offline vLLM prompt logprobs. Steps are delimited by paragraph breaks. Step confidence $c_i$ is the arithmetic mean of $p_\tau^{\max}$ over the step's token span, and volatility is $v_i=(c_i-c_{i-1})^2/4$ as in Section~\ref{sec:local_overthink_detection}. The four quantiles $(q_{25c},q_{75c},q_{25v},q_{75v})$ are calibrated once on the model's self-generated trace corpus.

\paragraph{Judge re-checking.}
Every confidence-screened \textsc{Overthink} step is re-checked by Qwen3-30B using vLLM. The judge sees the problem, the preceding steps, the candidate step, and the predicted label, and returns a corrected label plus a short rationale in JSON. The template is:
\begin{verbatim}
You are evaluating whether an existing overthink label
for one reasoning step is correct.

Problem:
{question}

Previous reasoning steps:
{previous_steps}

Current reasoning step:
{current_step}

Predicted label:
{predicted_label}

Label definitions:
- OVERTHINK: this step is wasteful or redundant;
  removing it would not break the path to the answer.
- NORMAL: this step is necessary reasoning;
  removing it would break or weaken the path.
- FINAL_ANSWER_WRAPUP: this step mainly wraps up
  or restates the final answer.

Respond in this exact JSON format and nothing else:
{
  "gold_label": "OVERTHINK|NORMAL|FINAL_ANSWER_WRAPUP",
  "reason": "short reason"
}
\end{verbatim}

Only judge-confirmed \textsc{overthink} steps survive. A trace with many flagged steps could otherwise dominate regeneration, so each trace keeps at most 10 candidate states, selected by uniform sampling.

\subsection{Evaluation Detail}
\label{app:evaluation}

\paragraph{Evaluation prompts.}
Each benchmark appends a short instruction to the raw problem text, and the result is wrapped by each model's native chat template. The five mathematical benchmarks share the instruction
\begin{quote}
\small\texttt{Please reason step by step, and put your final answer within \textbackslash boxed\{\}.}
\end{quote}
The three code benchmarks share the following instruction:
\begin{quote}
\small\texttt{Please write a solution in Python. Wrap your final code in a \textasciigrave\textasciigrave\textasciigrave python code block.}
\end{quote}
Following their official protocols, MBPP additionally includes the public tests in the prompt (\texttt{Your code should pass these tests: {tests}}), while LiveCodeBench appends the provided starter code when applicable. We extract the last Python code block from the complete output, which naturally excludes any code blocks generated within the thinking process, and execute it in a sandbox with per-problem timeouts of 10 seconds for MBPP and 6 seconds for LiveCodeBench. CodeContests follows the AlphaCode judging protocol, where outputs are split by whitespace and compared token-wise case-insensitively, with exact matching for integers and a tolerance of $10^{-5}$ for floating-point values.

\subsection{Ablation Construction}
\label{app:ablation_details}

This subsection provides the full descriptions of the ablation variants introduced in the main text. All variants use the same question set, pair-selection thresholds, and training recipe as LoCo. Each variant modifies only one construction choice, while differences in pair counts arise naturally from the resulting construction.

\paragraph{Global.}
The Global variant pairs the shortest and longest correct rollouts of the same question. To avoid disadvantaging this variant, its pool is expanded from three source rollouts to eight per question, allowing the selected responses to draw from a larger sample of short solutions than the source traces provide. The same savings threshold and top-$K$ selection as LoCo are applied, and the prompt contains only the question, without a frozen state.

\paragraph{GPT-5.4 Cont.}
GPT-5.4 continues from the same frozen prefix, provided as an assistant prefill with a system instruction to continue the reasoning in the same first-person style without restarting the problem. The continuation is generated directly from the frozen prefix and replaces the self-generated chosen continuation only when it is correct and shorter than the rejected continuation. Thus, this variant differs from LoCo only on the chosen side.

\paragraph{RandCut, UniCut, and PosMat.}
Each trace receives the same number of cutpoints as selected by the detector. Cutpoints are sampled from all eligible steps after excluding the first step and the final two steps. RandCut samples uniformly at random. UniCut places the cutpoints on an evenly spaced grid over the trace and snaps each to the nearest eligible step in its segment. 
PosMat samples cutpoints randomly according to the detector's relative-position distribution, matching its cutpoint locations without the confidence signal for selection. 
Downstream regeneration and pair construction remain unchanged. The detector's cutpoints concentrate late in the trace, with a mean relative position of 0.57 and about half occurring in the final 40\% of steps. In contrast, RandCut and UniCut cover the trace more uniformly, with mean relative positions of 0.47 and 0.50, respectively. PosMat matches the detector's 0.57 by construction, so the remaining gap between PosMat and LoCo isolates what the detector chooses to cut rather than where it cuts.

\section{Qualitative Examples}
\label{app:qualitative}

\subsection{Detection Traces}
\label{app:detection_traces}

\begin{figure}[t]
  \centering
  \begin{minipage}[t]{\linewidth}
    \centering
    \includegraphics[width=\linewidth]{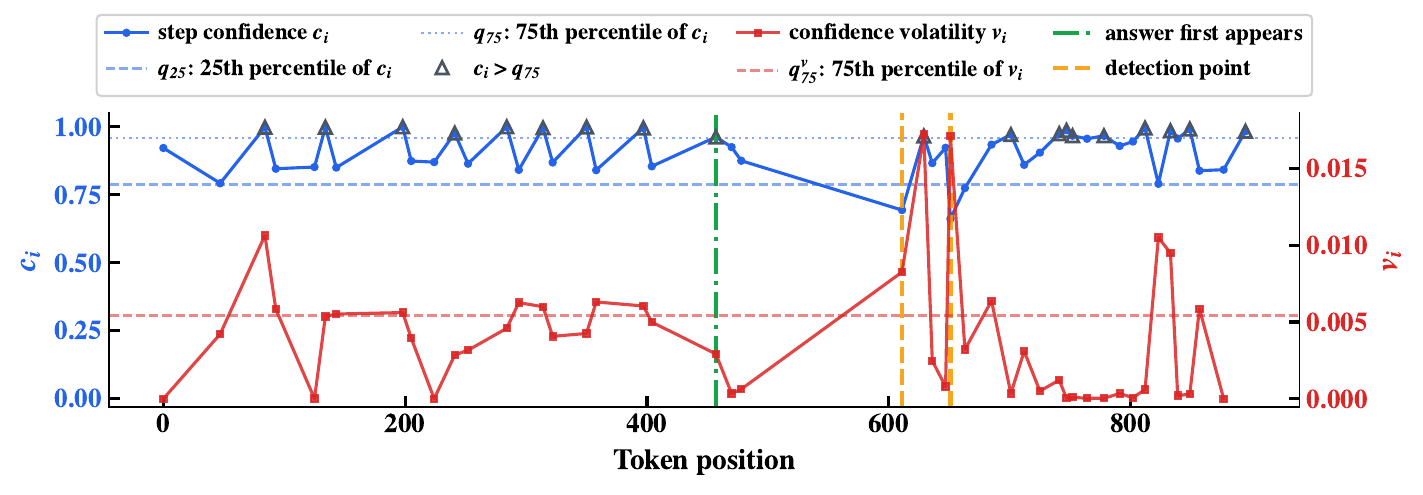}
    {\footnotesize\textbf{Trace (a)}}
  \end{minipage}
  \vspace{0.6em}
  \begin{minipage}[t]{\linewidth}
    \centering
    \includegraphics[width=\linewidth]{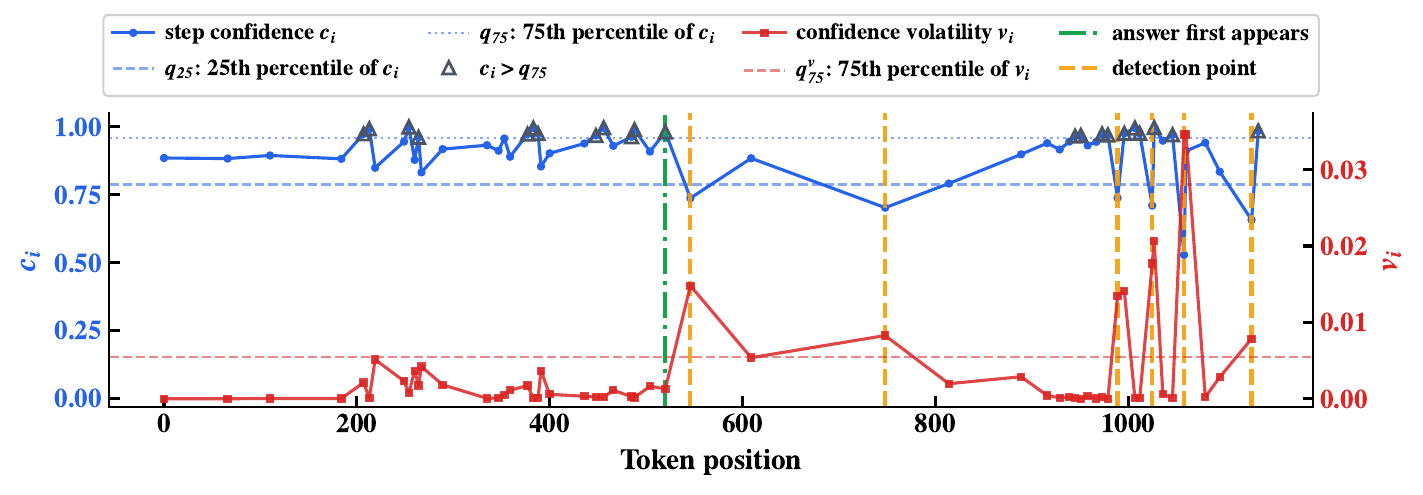}
    {\footnotesize\textbf{Trace (b)}}
  \end{minipage}
  \caption{Step confidence $c_i$ (blue) and confidence volatility $v_i$ (red) along two representative reasoning traces. The green dash-dotted line indicates the first occurrence of the correct answer. Orange dashed lines mark detected overthinking states ($c_i\le q_{25}$ and $v_i\ge q_{75}^v$), while triangles indicate high-confidence steps ($c_i>q_{75}$).}
  \label{fig:detection_traces}
\end{figure}

Figure~\ref{fig:detection_traces} shows the step confidence $c_i$ and volatility $v_i$ on two representative traces from the training data, together with the thresholds and detected states. The triangles highlight high-confidence steps, illustrating that using confidence alone would flag many additional steps as potential cutpoints. By jointly considering confidence and volatility, our detector substantially reduces such false positives. 

The full traces are shown, with green marking the step where the correct answer first appears, orange marking the detected overthinking states, and $\blacktriangle$ marking high-confidence steps.

\paragraph{Trace (a)}

\textbf{Question:} If $x = 101$ and $x^3y - 2x^2y + xy = 101000$, what is the value of $y$?

The answer $\frac{1}{10}$ first appears at step 20, and the detector fires at steps 23 and 27. The 48 steps are:

\begin{quote}\small
\textbf{Step 0.} Alright, so I have this equation: \( x^3 y - 2x^2 y + x y = 101000 \), and I know that \( x = 101 \). I need to find the value of \( y \). Hmm, okay. Let me think about how to approach this.

\textbf{Step 1.} First off, the equation looks a bit complicated with all those terms, but maybe I can factor it or simplify it somehow. Let me write it down again:

\textbf{Step 2.} $\blacktriangle$ \( x^3 y - 2x^2 y + x y = 101000 \).

\textbf{Step 3.} I notice that each term on the left side has a factor of \( x y \). That seems promising. Maybe I can factor that out. Let me try:

\textbf{Step 4.} Factor out \( x y \) from each term:

\textbf{Step 5.} $\blacktriangle$ \( x y (x^2 - 2x + 1) = 101000 \).

\textbf{Step 6.} Wait, that simplifies nicely. Now, inside the parentheses, \( x^2 - 2x + 1 \) is a quadratic expression. I remember that this is a perfect square because \( (x - 1)^2 = x^2 - 2x + 1 \). So, I can rewrite the equation as:

\textbf{Step 7.} $\blacktriangle$ \( x y (x - 1)^2 = 101000 \).

\textbf{Step 8.} Great, that's simpler. So now, the equation is \( x y (x - 1)^2 = 101000 \).

\textbf{Step 9.} Given that \( x = 101 \), let me substitute that into the equation:

\textbf{Step 10.} $\blacktriangle$ \( 101 \cdot y \cdot (101 - 1)^2 = 101000 \).

\textbf{Step 11.} Let me compute \( (101 - 1) \) first. That's \( 100 \). Then, squaring that gives \( 100^2 = 10000 \). So, plugging that back in:

\textbf{Step 12.} $\blacktriangle$ \( 101 \cdot y \cdot 10000 = 101000 \).

\textbf{Step 13.} Hmm, okay. So, \( 101 \times 10000 = 1010000 \). Therefore, the equation becomes:

\textbf{Step 14.} $\blacktriangle$ \( 1010000 \cdot y = 101000 \).

\textbf{Step 15.} Now, to solve for \( y \), I need to divide both sides of the equation by \( 1010000 \). Let me write that down:

\textbf{Step 16.} $\blacktriangle$ \( y = \frac{101000}{1010000} \).

\textbf{Step 17.} Simplify this fraction. Both numerator and denominator have the same number of zeros, so I can divide numerator and denominator by 1000 to make it easier:

\textbf{Step 18.} $\blacktriangle$ \( y = \frac{101}{1010} \).

\textbf{Step 19.} Hmm, 101 is a prime number, right? So, I can't simplify this fraction any further unless 101 divides into 1010, which it does. Let me check: 1010 divided by 101 is 10, because 101 times 10 is 1010. So, simplifying:

\textcolor{green}{\textbf{Step 20.}} \( y = \frac{101}{101 \times 10} = \frac{1}{10} \).

\textbf{Step 21.} So, \( y = \frac{1}{10} \) or 0.1.

\textbf{Step 22.} Wait a second, let me double-check my steps to make sure I didn't make a mistake. Factoring out \( x y \) from each term was correct, and recognizing \( x^2 - 2x + 1 \) as \( (x - 1)^2 \) is right. Then substituting \( x = 101 \), so \( (101 - 1)^2 = 100^2 = 10000 \), that's correct. Multiplying \( 101 \times 10000 \) gives 1010000, which is correct. Then setting up the equation \( 1010000 y = 101000 \), so dividing both sides by 1010000 gives \( y = 101000 / 1010000 \), which simplifies to \( 1/10 \). Yep, that seems right.

\textcolor{orange}{\textbf{Step 23.}} Alternatively, I can think of it as moving the 1010000 to the right side:

\textbf{Step 24.} $\blacktriangle$ \( y = 101000 / 1010000 \).

\textbf{Step 25.} Dividing numerator and denominator by 101000:

\textbf{Step 26.} \( y = 1 / 10 \).

\textcolor{orange}{\textbf{Step 27.}} Yep, same result. So, I think that's correct.

\textbf{Step 28.} Just to make sure, maybe I can plug the value back into the original equation to verify.

\textbf{Step 29.} Given \( x = 101 \) and \( y = 1/10 \), let's compute each term:

\textbf{Step 30.} $\blacktriangle$ First term: \( x^3 y = 101^3 \times (1/10) \).

\textbf{Step 31.} 101 cubed is 101 * 101 * 101. Let me compute that.

\textbf{Step 32.} First, 101 * 101 is 10201. Then, 10201 * 101. Let's compute that:

\textbf{Step 33.} $\blacktriangle$ 10201 * 100 = 1,020,100

\textbf{Step 34.} $\blacktriangle$ Plus 10201 * 1 = 10,201

\textbf{Step 35.} $\blacktriangle$ Adding together: 1,020,100 + 10,201 = 1,030,301.

\textbf{Step 36.} So, \( x^3 y = 1,030,301 \times (1/10) = 103,030.1 \).

\textbf{Step 37.} $\blacktriangle$ Second term: \( -2x^2 y = -2 * (101)^2 * (1/10) \).

\textbf{Step 38.} 101 squared is 10,201. So, 2 * 10,201 = 20,402.

\textbf{Step 39.} Then, \( -20,402 * (1/10) = -2,040.2 \).

\textbf{Step 40.} $\blacktriangle$ Third term: \( x y = 101 * (1/10) = 10.1 \).

\textbf{Step 41.} Now, adding all three terms together:

\textbf{Step 42.} $\blacktriangle$ 103,030.1 - 2,040.2 + 10.1.

\textbf{Step 43.} First, 103,030.1 - 2,040.2 = 100,989.9.

\textbf{Step 44.} $\blacktriangle$ Then, 100,989.9 + 10.1 = 101,000.

\textbf{Step 45.} Which matches the right side of the equation, 101,000. Perfect, that checks out.

\textbf{Step 46.} So, I can confidently say that \( y = 1/10 \) is the correct solution.

\textbf{Step 47.} \textbf{Final Answer}
The value of \( y \) is \boxed{\dfrac{1}{10}}.
\texttt{</think>}

\end{quote}

\paragraph{Trace (b)}

\textbf{Question:} What is the quotient when $8x^3+16x^2-7x+4$ is divided by $2x+5$?

The answer $4x^2 -2x + \frac{3}{2}$ first appears at step 29, and the detector fires at steps 30, 32, 44, 48, 52, and 56. The 58 steps are:

\begin{quote}\small
\textbf{Step 0.} Okay, so I need to find the quotient when dividing the polynomial \(8x^3 + 16x^2 -7x +4\) by \(2x +5\). Hmm, I remember that polynomial division is kind of like long division with numbers, but with polynomials. Let me think about how to approach this step by step.

\textbf{Step 1.} First, I should probably set it up like a long division problem. So, the dividend is \(8x^3 + 16x^2 -7x +4\) and the divisor is \(2x +5\). I need to divide each term step by step.

\textbf{Step 2.} Let me recall the process. I need to divide the first term of the dividend by the first term of the divisor. So, the first term of the dividend is \(8x^3\) and the first term of the divisor is \(2x\). Dividing \(8x^3\) by \(2x\) gives me \(4x^2\). That should be the first term of the quotient.

\textbf{Step 3.} Now, I need to multiply the entire divisor \(2x +5\) by this term \(4x^2\). Let me do that:

\textbf{Step 4.} $\blacktriangle$ \(4x^2 \times 2x = 8x^3\)

\textbf{Step 5.} $\blacktriangle$ \(4x^2 \times 5 = 20x^2\)

\textbf{Step 6.} So, multiplying gives me \(8x^3 +20x^2\). Next, I subtract this from the original dividend. So, subtracting term by term:

\textbf{Step 7.} \(8x^3 -8x^3 = 0\)

\textbf{Step 8.} $\blacktriangle$ \(16x^2 -20x^2 = -4x^2\)

\textbf{Step 9.} \(-7x -0 = -7x\)

\textbf{Step 10.} $\blacktriangle$ \(4 -0 =4\)

\textbf{Step 11.} So, after subtracting, the new polynomial is \(-4x^2 -7x +4\). That's our new dividend now.

\textbf{Step 12.} I repeat the process. Divide the first term of the new dividend \(-4x^2\) by the first term of the divisor \(2x\). That gives me \(-2x\). So, the next term in the quotient is \(-2x\).

\textbf{Step 13.} Now, multiply the divisor \(2x +5\) by \(-2x\):

\textbf{Step 14.} \( -2x \times 2x = -4x^2\)

\textbf{Step 15.} \(-2x \times 5 = -10x\)

\textbf{Step 16.} So, that gives \(-4x^2 -10x\). Subtract this from the current dividend:

\textbf{Step 17.} $\blacktriangle$ \(-4x^2 - (-4x^2) = 0\)

\textbf{Step 18.} $\blacktriangle$ \(-7x - (-10x) = 3x\)

\textbf{Step 19.} $\blacktriangle$ \(4 -0 =4\)

\textbf{Step 20.} So, the new polynomial is \(3x +4\).

\textbf{Step 21.} Again, I divide the first term \(3x\) by \(2x\), which gives me \(\frac{3}{2}\) or \(1.5\). So, the next term in the quotient is \(\frac{3}{2}\).

\textbf{Step 22.} Multiply the divisor \(2x +5\) by \(\frac{3}{2}\):

\textbf{Step 23.} $\blacktriangle$ \(2x \times \frac{3}{2} = 3x\)

\textbf{Step 24.} $\blacktriangle$ \(5 \times \frac{3}{2} = \frac{15}{2}\)

\textbf{Step 25.} So, that gives \(3x + \frac{15}{2}\). Subtract this from the current dividend:

\textbf{Step 26.} $\blacktriangle$ \(3x -3x =0\)

\textbf{Step 27.} $\blacktriangle$ \(4 - \frac{15}{2} = \frac{8}{2} - \frac{15}{2} = -\frac{7}{2}\)

\textbf{Step 28.} So, after subtracting, I have a remainder of \(-\frac{7}{2}\).

\textcolor{green}{\textbf{Step 29.}} Putting it all together, the quotient is \(4x^2 -2x + \frac{3}{2}\) with a remainder of \(-\frac{7}{2}\).

\textcolor{orange}{\textbf{Step 30.}} Wait, but I need to write this as a quotient, so do I include the remainder? The problem says "the quotient when...", which usually implies just the polynomial part, not including the remainder. So, maybe I should just write \(4x^2 -2x + \frac{3}{2}\).

\textbf{Step 31.} Let me double-check my work to make sure I didn't make any mistakes. So, starting with dividing \(8x^3\) by \(2x\), that's \(4x^2\). Multiply back \(4x^2(2x +5)\) gives \(8x^3 +20x^2\). Subtracting that from the original gives \(-4x^2 -7x +4\). Then, dividing \(-4x^2\) by \(2x\) is \(-2x\). Multiply back \(-2x(2x +5)\) gives \(-4x^2 -10x\). Subtracting that gives \(3x +4\). Then, dividing \(3x\) by \(2x\) is \(1.5\). Multiply back \(1.5(2x +5)\) gives \(3x +7.5\). Subtracting that gives \(4 -7.5 = -3.5\) or \(-\frac{7}{2}\). So, that seems correct.

\textcolor{orange}{\textbf{Step 32.}} Alternatively, maybe I can express the quotient as a polynomial with fractions. So, \(4x^2 -2x + \frac{3}{2}\). Alternatively, I can write it with all terms over 2 to make it a single fraction, but I think the problem just wants the quotient, so probably that's fine.

\textbf{Step 33.} Alternatively, let me think if I can write the quotient as a mixed expression or if there's another way. But in polynomial division, the quotient is just the polynomial part, and the remainder is separate. So, yeah, \(4x^2 -2x + \frac{3}{2}\) is the quotient, and \(-\frac{7}{2}\) is the remainder.

\textbf{Step 34.} Wait, let me check if \( (2x +5)(4x^2 -2x + \frac{3}{2}) + (-\frac{7}{2}) \) equals the original polynomial.

\textbf{Step 35.} Let me compute \( (2x +5)(4x^2 -2x + \frac{3}{2}) \):

\textbf{Step 36.} First, multiply \(2x\) with each term:

\textbf{Step 37.} \(2x \times 4x^2 =8x^3\)

\textbf{Step 38.} $\blacktriangle$ \(2x \times (-2x) =-4x^2\)

\textbf{Step 39.} $\blacktriangle$ \(2x \times \frac{3}{2} =3x\)

\textbf{Step 40.} Then, multiply \(5\) with each term:

\textbf{Step 41.} \(5 \times4x^2 =20x^2\)

\textbf{Step 42.} $\blacktriangle$ \(5 \times (-2x) =-10x\)

\textbf{Step 43.} $\blacktriangle$ \(5 \times \frac{3}{2} = \frac{15}{2}\)

\textcolor{orange}{\textbf{Step 44.}} Now, add all these together:

\textbf{Step 45.} $\blacktriangle$ \(8x^3 -4x^2 +3x +20x^2 -10x + \frac{15}{2}\)

\textbf{Step 46.} $\blacktriangle$ Combine like terms:

\textbf{Step 47.} $\blacktriangle$ \(8x^3 + (-4x^2 +20x^2) + (3x -10x) + \frac{15}{2}\)

\textcolor{orange}{\textbf{Step 48.}} Simplify:

\textbf{Step 49.} $\blacktriangle$ \(8x^3 +16x^2 -7x + \frac{15}{2}\)

\textbf{Step 50.} Now, add the remainder \(-\frac{7}{2}\):

\textbf{Step 51.} $\blacktriangle$ \(8x^3 +16x^2 -7x + \frac{15}{2} - \frac{7}{2}\)

\textcolor{orange}{\textbf{Step 52.}} Which is:

\textbf{Step 53.} \(8x^3 +16x^2 -7x +4\), which matches the original dividend. So, that checks out.

\textbf{Step 54.} Therefore, the quotient is indeed \(4x^2 -2x + \frac{3}{2}\).

\textbf{Step 55.} Alternatively, I can write \(\frac{3}{2}\) as \(1.5\), but in fraction form, it's probably better to leave it as \(\frac{3}{2}\).

\textcolor{orange}{\textbf{Step 56.}} So, I think that's the answer.

\textbf{Step 57.} \textbf{Final Answer}
The quotient is \boxed{4x^2 - 2x + \dfrac{3}{2}}.
\texttt{</think>}

\end{quote}

\subsection{Preference Pairs}
\label{app:preference_pairs}

We present preference triples from the training data in full. Each triple contains the frozen prefix $P$, the chosen continuation $y^{+}$, which is the shortest verified regeneration from the same state, and the rejected original continuation $y^{-}$, reproduced verbatim from the stored pairs. The model's LaTeX and Markdown are rendered as-is, with only leading and trailing whitespace trimmed. Each continuation ends its reasoning block with the closing \texttt{</think>} tag, followed by the final answer block outside the reasoning region.

\paragraph{Example 1}

\textbf{Question:} If July 3, 2030 is a Wednesday, what day of the week is July 14, 2030?

The detector flags the state 28 steps into the original trace. By this point, the answer (Sunday) has already been derived arithmetically, confirmed by a day-by-day listing, and re-confirmed using modular arithmetic. The preference pair contrasts the model's original continuation from this state, which is rejected, with the shortest verified regeneration, which is chosen.

\paragraph{Frozen prefix $P$ (the first 28 steps).}
\begin{quote}\small
Okay, so I need to figure out what day of the week July 14, 2030 is, given that July 3, 2030 is a Wednesday. Hmm, let's see. I remember that each week has 7 days, so the days of the week repeat every 7 days. That means if I can find out how many days are between July 3 and July 14, I can add that number of days to Wednesday and find the corresponding day.

First, let me count the number of days between July 3 and July 14. Let's see, July has 31 days, right? So from July 3 to July 14, how many days is that? Hmm, if I subtract 3 from 14, that would give me 11 days, but wait, does that include both the start and end dates? I think not. So if I start counting from July 3 as day 0, then July 4 is day 1, and so on. So to get from July 3 to July 14, I need to count 11 days. Let me verify that: July 3, 4, 5, 6, 7, 8, 9, 10, 11, 12, 13, 14. Yeah, that's 11 days. So the difference is 11 days.

Wait, but sometimes when calculating the number of days between two dates, you have to be careful whether to include both the start and end dates. Let me think. If I have two consecutive days, like July 3 and July 4, that's 1 day difference, right? So subtracting 3 from 4 gives 1, which is correct. So in this case, July 3 to July 14 is 11 days. So 14 minus 3 is 11, so that's correct.

Now, since each week is 7 days, I can figure out how many weeks and extra days are in 11 days. So, 11 divided by 7 is 1 week and 4 days. So 11 days is equal to 1 week plus 4 days. That means that 11 days from a Wednesday would be Wednesday plus 4 days. Let me count: Wednesday plus 1 day is Thursday, plus 2 is Friday, plus 3 is Saturday, plus 4 is Sunday. So, is July 14, 2030 a Sunday?

Wait, let me double-check that. Sometimes when I'm adding days, I can get confused. Maybe I should list out the dates and their corresponding days.

Starting from July 3, which is Wednesday. Let's list each day:

July 3: Wednesday

July 4: Thursday

July 5: Friday

July 6: Saturday

July 7: Sunday

July 8: Monday

July 9: Tuesday

July 10: Wednesday

July 11: Thursday

July 12: Friday

July 13: Saturday

July 14: Sunday

Wait, so July 14 is indeed a Sunday. Hmm, okay, so my initial calculation was correct. 11 days from Wednesday is Sunday.

But hold on, sometimes people might get confused whether to add 11 days or 12 days because of the way we count dates. Let me see: If July 3 is day 0, then July 14 is day 11. So adding 11 days to Wednesday. 7 days later is the same day of the week, which would be Wednesday again, and then 4 more days would bring it to Sunday.

Alternatively, another way to think about it is using modular arithmetic. Since the days cycle every 7 days, the day of the week repeats every 7 days. So, if I have a certain number of days, I can find the remainder when divided by 7, and that will tell me how many days ahead it is from the starting day.

So, 11 divided by 7 is 1 with a remainder of 4. So, the day of the week would be Wednesday plus 4 days, which is Sunday. So that matches my previous result.

But just to be thorough, let me check another method. Maybe using the concept of odd days. Each month, the number of days can affect the day of the week for the next month. Wait, but in this case, we're just dealing with the same month, July. So maybe that's complicating things. Alternatively, I can use the concept of counting weeks and extra days.

Wait, another way: If July 3 is Wednesday, then July 10 would also be a Wednesday because that's exactly 7 days later. Then, July 10 is Wednesday, so July 11 is Thursday, July 12 is Friday, July 13 is Saturday, July 14 is Sunday. So that also gets me to Sunday. Yeah, that seems consistent.

Alternatively, I could use a reference point. I know that July 4, 2023, is a Monday. Wait, no, that's not relevant. Maybe I can find a known date in 2030. Wait, 2030 is not a leap year, so February has 28 days. But that might not help directly with July.

Alternatively, maybe I can use Zeller's Congruence or some algorithm to calculate the day of the week for July 14, 2030, given that July 3 is a Wednesday. But that might be overcomplicating, since I already have two methods confirming it's Sunday.

Wait, let me also just consider the number of weeks and days. From July 3 to July 14 is 11 days. 11 divided by 7 is 1 week and 4 days. So, starting from Wednesday, adding 4 days: Wednesday + 4 days = Sunday.

Alternatively, if I consider that 7 days later is Wednesday, then adding another 4 days: Thursday, Friday, Saturday, Sunday. So that also gives me Sunday.
\end{quote}

\paragraph{Chosen $y^{+}$ (212 tokens).}
\begin{quote}\small
So, multiple methods confirm that July 14, 2030 is a Sunday.

I think I'm confident now that the answer is Sunday.

**Final Answer**
\boxed{Sunday}
\texttt{</think>}

Given that July 3, 2030 is a Wednesday, we need to determine the day of the week for July 14, 2030.

First, we calculate the number of days between July 3 and July 14. This is 11 days. 

Next, we determine how many weeks and extra days are in 11 days. Dividing 11 by 7 gives 1 week and 4 extra days. 

Starting from Wednesday, we add 4 days:
- Wednesday + 1 day = Thursday
- Thursday + 1 day = Friday
- Friday + 1 day = Saturday
- Saturday + 1 day = Sunday

Thus, July 14, 2030 is a Sunday.

\[
\boxed{Sunday}
\]
\end{quote}

\paragraph{Rejected $y^{-}$ (986 tokens, pair gap 774).}
\begin{quote}\small
Wait, another way: Maybe using the concept of reference dates. For example, I know that January 1, 2030, was a Wednesday. Wait, no, actually, let me check. Wait, no, I don't know that off the top of my head. Maybe it's better not to go that route.

Alternatively, I can use the fact that each year, the same date moves one day forward in the week, unless it's a leap year, in which case it moves two days. Wait, 2030 is not a leap year, so the next year, 2031, same date would be one day later in the week. But that might not help here.

Wait, perhaps I can count the number of days between July 3, 2030, and July 14, 2030, and then see how many weeks and extra days that is.

Wait, so July 3 to July 14 is 11 days, as established earlier. So 11 days later. 11 divided by 7 is 1 week and 4 days. So, Wednesday plus 4 days is Sunday.

Alternatively, maybe using the concept of counting weeks: 7 days is one week, so 14 days would be two weeks, which brings us back to Wednesday. But 11 days is less than 14, so it's 1 week and 4 days.

Alternatively, perhaps thinking in terms of modulo 7 arithmetic. If I assign numbers to days: Sunday=0, Monday=1, Tuesday=2, Wednesday=3, Thursday=4, Friday=5, Saturday=6. Then, adding 11 days to Wednesday (which is 3) would be 3 + 11 = 14. Then, 14 mod 7 is 0, which is Sunday. So that also works.

Wait, so 3 (Wednesday) + 11 days = 14, which is 0 mod 7, which is Sunday. So that confirms it as well.

Alternatively, if I use the formula for the day of the week: (start\_day + number\_of\_days) mod 7. So, if start\_day is Wednesday, which is 3, number\_of\_days is 11, so (3 + 11) mod 7 = 14 mod 7 = 0, which is Sunday.

Alternatively, if I use another numbering system where Sunday is 1, Monday is 2, etc., but that might complicate things. The key is that modulo 7 will give the correct offset.

So, given that, I think I'm confident that July 14, 2030 is a Sunday.

But just to be absolutely sure, let me recount the days manually:

Starting from July 3, which is Wednesday.

July 3: Wed

July 4: Thu

July 5: Fri

July 6: Sat

July 7: Sun

July 8: Mon

July 9: Tue

July 10: Wed

July 11: Thu

July 12: Fri

July 13: Sat

July 14: Sun

Yes, so each week, it cycles back. So from July 3 to July 10 is exactly one week, which is 7 days, so that brings us back to Wednesday. Then, July 11 is Thursday, 12 is Friday, 13 is Saturday, and 14 is Sunday. So that's four days after Wednesday, which is Sunday.

Therefore, I can conclude that July 14, 2030 is a Sunday.

**Final Answer**
\boxed{Sunday}
\texttt{</think>}

Given that July 3, 2030 is a Wednesday, we need to determine the day of the week for July 14, 2030.

First, we calculate the number of days between July 3 and July 14, 2030. This is 11 days (from July 3 to July 14 inclusive).

Next, we determine how many weeks and extra days are in 11 days:
- 11 days is equivalent to 1 week (7 days) and 4 extra days.

Starting from Wednesday, we add 4 days:
- Wednesday + 1 day = Thursday
- Thursday + 1 day = Friday
- Friday + 1 day = Saturday
- Saturday + 1 day = Sunday

Thus, 11 days after Wednesday is Sunday.

Therefore, July 14, 2030 is a Sunday.

\[
\boxed{Sunday}
\]
\end{quote}

The rejected suffix adds no new information once the answer has been derived and confirmed. It restarts from a reference-date idea, abandons it, re-derives the offset through counting, modulo-7 arithmetic, and a day-of-week formula, and finally recounts the calendar by hand before restating Sunday. At this state, LoCo learns a simple preference to conclude rather than continue.

\paragraph{Example 2}

\textbf{Question:} What fraction of $\frac{1}{2}$ is $\frac{1}{7}$?

The detector flags the state 11 steps into the trace, after the equation $x \cdot \frac{1}{2} = \frac{1}{7}$ has been set up, solved, and verified twice.

\paragraph{Frozen prefix $P$ (the first 11 steps).}
\begin{quote}\small
Alright, so I have this math problem here: "What fraction of 1/2 is 1/7?" Hmm, okay. I need to figure out what fraction, when multiplied by 1/2, gives me 1/7. Let me think about how to approach this.

First, let's translate the question into an equation. If I let the unknown fraction be x, then the problem is asking: x times 1/2 equals 1/7. So, mathematically, that would be:

x * (1/2) = 1/7

Now, to solve for x, I need to isolate it. Since it's multiplied by 1/2, I can do the opposite operation, which is multiplication by the reciprocal. The reciprocal of 1/2 is 2/1, or just 2. So, if I multiply both sides of the equation by 2, I should get:

x = (1/7) * 2

Let me compute that. 1/7 multiplied by 2 is 2/7. So, x equals 2/7. That means 2/7 of 1/2 is 1/7.

Wait, let me double-check that. If I take 2/7 and multiply it by 1/2, does that give me 1/7?

Calculating that: (2/7) * (1/2) = (2*1)/(7*2) = 2/14 = 1/7. Yep, that checks out.

But just to make sure I didn't make a mistake in setting up the equation, let me rephrase the question. It's asking, "What fraction of 1/2 is 1/7?" So, it's like saying 1/7 is what portion of 1/2? That portion is x.

Another way to think about it is: if 1/2 is the whole, then 1/7 is some fraction of that whole. So, x = (1/7) / (1/2). Dividing by a fraction is the same as multiplying by its reciprocal, so x = (1/7) * (2/1) = 2/7.

Yes, that makes sense. So, the fraction is 2/7.
\end{quote}

\paragraph{Chosen $y^{+}$ (296 tokens).}
\begin{quote}\small
I think that's the correct answer.

Just to recap, the steps are:

1. Set up the equation: x * (1/2) = 1/7.
2. Solve for x by multiplying both sides by 2: x = (1/7) * 2 = 2/7.
3. Verify by plugging back in: (2/7) * (1/2) = 1/7.

Everything seems to add up, so I'm confident the answer is 2/7.

**Final Answer**
The fraction is \boxed{\dfrac{2}{7}}.
\texttt{</think>}

To determine what fraction of \(\frac{1}{2}\) is \(\frac{1}{7}\), we set up the equation:

\[ x \times \frac{1}{2} = \frac{1}{7} \]

To solve for \(x\), we multiply both sides by the reciprocal of \(\frac{1}{2}\), which is 2:

\[ x = \frac{1}{7} \times 2 \]

This simplifies to:

\[ x = \frac{2}{7} \]

We verify this by checking that \(\frac{2}{7} \times \frac{1}{2} = \frac{1}{7}\), which is correct.

Thus, the fraction is \(\boxed{\dfrac{2}{7}}\).
\end{quote}

\paragraph{Rejected $y^{-}$ (1{,}247 tokens, pair gap 951).}
\begin{quote}\small
Wait, let me try another approach just to be thorough. Let's use fractions in words. If I have half of something, and I want to find out how much 1/7 is of that half, I can set up a proportion.

Let me denote the unknown fraction as x. So, x of 1/2 is 1/7. So, x * (1/2) = 1/7. Solving for x, as before, gives me x = (1/7) / (1/2) = (1/7) * (2/1) = 2/7.

Alternatively, if I visualize this, imagine a pizza cut into 2 equal slices. Each slice is 1/2 of the pizza. Now, if I take 1/7 of a slice, how much of the whole pizza is that? It would be 1/7 of 1/2, which is 1/14 of the whole pizza. But wait, that seems different.

Wait, hold on. If I have 1/2 of a pizza, and I take 1/7 of that, it's 1/7 * 1/2 = 1/14 of the whole pizza. But the question isn't asking for that. The question is, what fraction of the 1/2 is the 1/7. So, actually, it's like, if I have 1/2, and 1/7 is some part of that 1/2. So, it's like, 1/7 divided by 1/2, which is 2/7, as we had before.

I think my confusion came from misinterpreting the visualization. So, the key is, it's not 1/7 of the whole pizza, but rather 1/7 is what fraction of the half. So, it's 1/7 divided by 1/2, which is 2/7. So, 1/7 is 2/7 of 1/2.

Another way to think about it is using ratios. If I have two numbers, a and b, and I want to find what fraction a is of b, it's a/b. So, in this case, a is 1/7 and b is 1/2. So, the fraction is (1/7)/(1/2) = 2/7.

Let me think of another way to confirm. If I have 1/2, and I want to know how many times 1/7 fits into it, it's like dividing 1/2 by 1/7. So, that would be (1/2) / (1/7) = 7/2 = 3.5. But that's not directly helpful because the question isn't asking how many times 1/7 fits into 1/2, but rather what fraction of 1/2 is 1/7.

Wait, so if I have 1/2 and 1/7 is a fraction of that, then it's 1/7 divided by 1/2, which is 2/7. So, 1/7 is 2/7 of 1/2.

Alternatively, using decimal equivalents might help. Let me convert the fractions to decimals.

1/2 is 0.5, and 1/7 is approximately 0.142857. So, what fraction of 0.5 is 0.142857? So, 0.142857 divided by 0.5 is approximately 0.285714, which is roughly 2/7, since 2 divided by 7 is approximately 0.285714. So, that aligns with our earlier result.

Okay, so all these methods point to the same answer: 2/7.

Just to recap, the steps are:

1. Let the unknown fraction be x.
2. Set up the equation: x * (1/2) = 1/7.
3. Solve for x by multiplying both sides by the reciprocal of 1/2, which is 2.
4. x = (1/7) * 2 = 2/7.
5. Verify by plugging back into the original equation: (2/7) * (1/2) = 1/7, which is correct.
6. Cross-checked using different methods: visualizing, ratios, decimal conversion---all confirm that 2/7 is the correct answer.

So, after going through multiple approaches and verifying each step, I feel confident that the fraction we're looking for is 2/7.

**Final Answer**
The fraction is \boxed{\dfrac{2}{7}}.
\texttt{</think>}

To determine what fraction of \(\frac{1}{2}\) is \(\frac{1}{7}\), we set up the equation where \(x\) is the unknown fraction:

\[ x \times \frac{1}{2} = \frac{1}{7} \]

To solve for \(x\), we multiply both sides by the reciprocal of \(\frac{1}{2}\), which is 2:

\[ x = \frac{1}{7} \times 2 \]

This simplifies to:

\[ x = \frac{2}{7} \]

We verify this by checking that \(\frac{2}{7}\) multiplied by \(\frac{1}{2}\) equals \(\frac{1}{7}\):

\[ \left(\frac{2}{7}\right) \times \left(\frac{1}{2}\right) = \frac{2 \times 1}{7 \times 2} = \frac{2}{14} = \frac{1}{7} \]

Thus, the fraction is \(\boxed{\dfrac{2}{7}}\).
\end{quote}

The rejected suffix re-derives the same result through a proportion, a pizza-slice visualization with a detour caused by a misinterpretation, a ratio argument, a dead-end attempt to count how many times the quantity fits, and a decimal-conversion check. It then closes with a step-by-step recap, adding four redundant derivations after the answer had already been verified in the prefix. LoCo anchors the preference at the state where the existing solution is already sufficient.

\end{document}